\documentclass[lettersize,journal]{IEEEtran}
\usepackage{amsmath,amsfonts,amssymb}
\usepackage{algorithmic}
\usepackage{algorithm}
\usepackage{array}
\usepackage[caption=false,font=normalsize,labelfont=sf,textfont=sf]{subfig}
\usepackage{textcomp}
\usepackage{stfloats}
\usepackage{url}
\usepackage{verbatim}
\usepackage{graphicx}
\graphicspath{{fig/}}
\usepackage{xcolor}
\usepackage{cite}
\begin{document}

\title{MoE-Based Learned Inertial Odometry for Bicycle Localization}

\author{Hao~Qiao,
        Yan~Wang,
        Shuo~Yang,
        Xiaoyao~Yu,
        Jian~Kuang,
        and~Xiaoji~Niu
\thanks{This work was supported in part by the National Key Research and Development Program of China under Grant 2024YFB3909200. (Corresponding author: Yan Wang)}
\thanks{Hao Qiao and Xiaoyao Yu are with the School of Geodesy and Geomatics, Wuhan University, Wuhan, Hubei 430079, China.(e-mail: hmqiao@whu.edu.cn; yuxiaoyao@whu.edu.cn)}
\thanks{Yan Wang and Jian Kuang are with the GNSS Research Center, Wuhan University, Wuhan, Hubei 430079, China.(e-mail: wystephen@whu.edu.cn; kuang@whu.edu.cn)}
\thanks{Shuo Yang is with the Hubei Luojia Laboratory, Wuhan, Hubei 430079, China.(e-mail: seokkk@whu.edu.cn)}
\thanks{Xiaoji Niu is with the GNSS Research Center, Wuhan University, Wuhan 430079, China, he is also with the Hubei Luojia Laboratory, Wuhan, Hubei 430079, China.(e-mail:xjniu@whu.edu.cn)}}

\markboth{Journal of \LaTeX\ Class Files,~Vol.~14, No.~8, August~2021}%
{Qiao \MakeLowercase{\textit{et al.}}: MoE-Based Learned Inertial Odometry for Bicycle Localization}

\maketitle

\bstctlcite{IEEEexample:BSTcontrol}

\begin{abstract}
GNSS suffers from multipath errors in urban canyons, making reliable bicycle localization difficult. Hand-crafted inertial alternatives, such as cycling dead reckoning and nonholonomic constraints, fail to generalize across riders, postures, and road surfaces. This work implements learned inertial odometry for bicycle localization: a neural velocity predictor, trained on multirider, multi-bicycle, and multi-surface IMU data, is tightly coupled with an extended Kalman filter for three-dimensional pose estimation, replacing hand-crafted cycling models with a data-driven representation. To reduce computational overhead, the predictor is realized as a sparsely-gated Mixture of Experts (MoE) network with top-$K$ routing, trained under an alternating parameter-freezing scheme with a per-expert capacity constraint to encourage diverse and balanced specialization.The proposed model attains 0.333 m/s inference error, 9.49 m ATE, and 2.58 m RTE at only 28.76 M FLOPs. This corresponds to approximately $7\times$ and $9\times$  fewer FLOPs than the ResMLP and ResNet backbones used in LLIO and TLIO, while maintaining comparable or superior accuracy. The proposed MoE variants further rank first on both ATE and RTE across all evaluation splits, demonstrating strong generalization.

\end{abstract}


\section{Introduction}
\IEEEPARstart{O}{wing} to their environmental friendliness, flexibility, and convenience, bicycles are widely used in transportation, rehabilitation training, military operations, and sports. As these applications continue to expand, the demand for high-precision bicycle localization has grown rapidly. In particular, the recent proliferation of bike-sharing systems has further highlighted the importance of efficient localization for effective fleet management.

At present, bicycle localization relies mainly on the Global Navigation Satellite System (GNSS), which has been widely adopted in commercial cycling devices (e.g., Garmin Edge) owing to its all-weather availability, low cost, and high accuracy. However, in environments such as urban canyons, tunnels, and tree-covered roads, GNSS signals are vulnerable to blockage and multipath effects, resulting in severe degradation or even complete failure of positioning. Consequently, GNSS alone cannot provide stable and continuous localization in complex scenarios.

As a complementary solution, Inertial Navigation Systems (INS) offer self-contained localization without relying on external signals. Driven by the rapid development of Micro-Electro-Mechanical Systems (MEMS), low-cost IMUs have been widely integrated into smartphones and wearable devices. Nevertheless, INS is inherently prone to error accumulation, which is further aggravated by the substantial noise of low-cost IMUs, thereby requiring additional motion constraints or external aiding information. For bicycle localization, Cycling Dead Reckoning (CDR)~\cite{chang2015improved} draws on Pedestrian Dead Reckoning (PDR) by estimating forward velocity from pedaling cadence to provide dead-reckoning updates. Separately, inertial navigation with nonholonomic constraints (NHC)~\cite{NHC-bicycle}, inspired by vehicle dead reckoning (VDR), enforces zero lateral and vertical velocities to constrain IMU drift. However, both approaches rely heavily on accurate physical modeling and sensor placement conditions.

More recently, data-driven inertial odometry has demonstrated clear advantages over traditional model-based methods. IONet~\cite{chen2018ionet} was the first to employ an LSTM network to estimate relative displacement on the ground plane. RoNIN~\cite{herath2020ronin} fused acceleration, angular velocity, and magnetic field measurements to infer global orientation, and then applied deep architectures (ResNet, LSTM, and TCN) for velocity estimation. IDOL~\cite{sun2021idol} jointly learned orientation and position, achieving higher accuracy. TLIO~\cite{liu2020tlio} tightly coupled deep networks with a Kalman filter to enable 3D pose estimation in complex scenarios, while CTIN~\cite{rao2022ctin} leveraged self-attention to capture both local and global context, thereby enhancing feature representation. Collectively, these studies have demonstrated strong accuracy and robustness across diverse motion patterns, providing valuable insights for advancing bicycle inertial odometry.

Despite these advances, the computational cost of data-driven methods remains a major bottleneck for deployment on mobile platforms. For instance, TLIO attains high accuracy with ResNet-based feature extraction but incurs heavy inference overhead. To address this limitation, several lightweight architectures have been proposed. LLIO~\cite{wang2022llio} employs a compact ResMLP to improve efficiency without sacrificing accuracy, whereas ResMixer~\cite{lai2024resmixer} replaces convolutional layers with Mixer layers to reduce computational load while preserving feature interactions. More recently, sparsely-gated Mixture of Experts (MoE) frameworks~\cite{shazeer2017outrageously} have been widely adopted to reduce computational cost through routing and load-balancing mechanisms, offering a promising direction for further efficiency gains.

Motivated by these observations, this work presents a lightweight learned inertial odometry for bicycles based on an MoE model, which provides 3D dead reckoning with low computational cost. The main contributions are summarized as follows:
\begin{itemize}
    \item We propose a learned inertial odometry for bicycle localization, in which a neural velocity predictor is tightly coupled with an extended Kalman filter to replace traditional model-based methods. The proposed method delivers higher overall localization accuracy than the traditional NHC scheme, with particularly stronger robustness on unpaved road. To the best of our knowledge, this is the first data-driven bicycle localization method.
    \item We realize the predictor as a sparsely-gated Mixture of Experts (MoE) network with top-$K$ routing, which retains the capacity of large models at a fraction of their compute. MoE-$8$-$2$-$128$ consumes only $28.76$\,M FLOPs, roughly $7\times$ and $9\times$ less than the ResMLP and ResNet backbones of LLIO and TLIO at comparable or superior accuracy, making it well suited for mobile deployment.
    \item We collect a large-scale cycling dataset covering diverse riders, bicycles, and road conditions, and conduct extensive experiments including overall performance comparison, robustness and generalization analysis, and ablation studies, demonstrating the accuracy, efficiency, and practicality of the proposed method.
\end{itemize}

\section{Related Work}
The proposed approach lies at the intersection of bicycle localization and learned inertial odometry. This section first reviews model-based bicycle localization methods and analyzes their limitations on real-world cycling, and then surveys data-driven inertial odometry, with particular attention to Mixture of Experts (MoE) architectures that are most closely related to the proposed design.
\subsection{Model-based Bicycle Localization}
Research on bicycle sensing has long relied on kinematic and dynamic models to characterize bicycle motion. Early efforts are primarily oriented toward control- and safety-related applications, including cycling accident detection~\cite{watthanawisuth2012wireless}, rider behavior analysis~\cite{kooijman2009some}, traffic impact assessment~\cite{shunping2008quantitative}, and cycling performance evaluation~\cite{eisenman2010bikenet}. In these studies, high-precision localization is not the primary objective, and the underlying models are not designed for trajectory reconstruction.

Beyond these application-oriented studies, subsequent work explicitly targets bicycle state estimation by fusing onboard sensors within physics-based frameworks. Motion models have been employed to reconstruct trajectories under specific maneuvers such as jumping, where strong motion constraints simplify the estimation~\cite{sadi2013new}. To mitigate IMU drift, force sensors have been combined with IMUs to recover bicycle attitude~\cite{zhang2013dynamic}, while low-cost MEMS sensors have been exploited together with cycling-specific cues such as forward velocity, traveled distance, and pedaling cadence to maintain localization under degraded GNSS conditions~\cite{chang2013cycling}. To further improve robustness in outdoor scenarios, multi-sensor platforms that integrate GNSS, IMU, magnetometer, and barometer have also been developed~\cite{chang2018method}.

Building on this trend, more recent studies formulate bicycle localization as an inertial-navigation problem with cycling-specific motion constraints. Inspired by Pedestrian Dead Reckoning (PDR), Cycling Dead Reckoning (CDR)~\cite{chang2015improved} first estimates the 3-D misalignment between the device frame and the bicycle frame from pedaling-induced acceleration patterns, and then models forward velocity as a linear function of pedaling cadence and per-cycle travel distance as a constant, providing velocity- and position-domain updates that bound IMU drift. In parallel, inertial navigation systems (INS) with nonholonomic constraints (NHC)~\cite{NHC-bicycle} enforce the zero lateral and vertical velocities typical of two-wheeled motion to enable dead-reckoning localization.

While effective when the underlying assumptions hold, both approaches depend critically on accurate physical modeling. CDR requires GNSS-aided calibration of its cadence--velocity coefficients and per-cycle travel distance, and treats them as time-invariant despite their dependence on terrain slope, rider posture, gear changes, and road surface; its cycle-based update further breaks down during coasting or irregular pedaling, and its misalignment estimation relies on quasi-steady straight riding and on prescribed device mounting locations such as the trunk, thigh, or lower leg, which limits its applicability to handlebar- or frame-mounted devices common in shared-bike scenarios. NHC, in turn, is sensitive to suspension-induced lever-arm variations, IMU mounting errors, and vibrations on uneven surfaces that violate the no-side-slip assumption. These limitations make precise modeling difficult and degrade localization accuracy and robustness in real-world cycling, motivating a shift toward data-driven solutions that learn bicycle-specific motion patterns directly from sensor measurements.

\subsection{Data-driven Inertial Odometry}
To overcome the difficulty of explicitly modeling MEMS sensor characteristics, recent work has increasingly adopted machine learning to learn motion representations directly from raw IMU data. Most of this progress has been concentrated on pedestrian and vehicle localization, where both network architectures and learning–filtering frameworks have evolved rapidly.

A representative early example is IONet~\cite{chen2018ionet}, which reformulates inertial odometry as a window-based displacement regression task: IMU measurements are segmented into fixed-length windows, an LSTM predicts the per-window displacement, and the trajectory is reconstructed by concatenating the predictions. Subsequent work extends this paradigm by jointly modeling pose and uncertainty, and by tightly coupling learning with classical filtering. TLIO~\cite{liu2020tlio} is a typical instance, in which a neural network predicts the displacement and its covariance and feeds them into an extended Kalman filter that recovers the full navigation state, including position, velocity, orientation, and IMU biases.

Beyond LSTM-based regression, a variety of neural architectures have been explored to improve representation capacity and efficiency. Examples include dedicated CNN backbones such as IMUNet~\cite{zeinali2024imunet} and Transformer-based models such as CTIN~\cite{rao2022ctin}. To enable on-device inference, lightweight networks such as LLIO-Net~\cite{wang2022llio} have been proposed to pursue a favorable accuracy--efficiency trade-off. These methods collectively demonstrate that learned representations can outperform hand-crafted models on common pedestrian and vehicle benchmarks.

More recently, the MoE model has also been explored for learned inertial odometry. M2EIT~\cite{li2025m2eit}densely fuses heterogeneous experts, specifically a residual network, a state-space model and a wavelet transform, via a multi-representation router, incurring computational cost that scales with the number of experts. In contrast, the proposed method adopts a sparsely-gated MoE architecture composed of homogeneous experts. These experts share a lightweight backbone yet specialize in distinct cycling motion patterns, with the primary goal of achieving low inference overhead for mobile deployment rather than maximizing representational richness.

Overall, although learned inertial odometry has been extensively investigated for pedestrian and vehicle localization and has achieved significant progress in efficiency, robustness, and generalization, these methods have rarely been tailored to the distinctive motion characteristics of bicycles, such as periodic vibrations, varying body postures, and frequent maneuvers. As a result, effective learned inertial localization for bicycles remains largely unexplored, which directly motivates the cycling-oriented sparsely-gated MoE architecture proposed in this paper.

\section{Proposed Framework}
\label{sec:system_overview}

The goal of this work is to achieve accurate and efficient three-dimensional bicycle localization using low-cost inertial sensors. The proposed framework is organized into four sequential stages: data collection, data processing, neural network training and filter inference. The first two stages constitute the offline data preparation pipeline, which produces the training-ready dataset, while the latter two stages form the learning and inference pipeline that delivers the final state estimates.

\begin{figure}[!t]
\centering
\includegraphics[width=0.8\columnwidth]{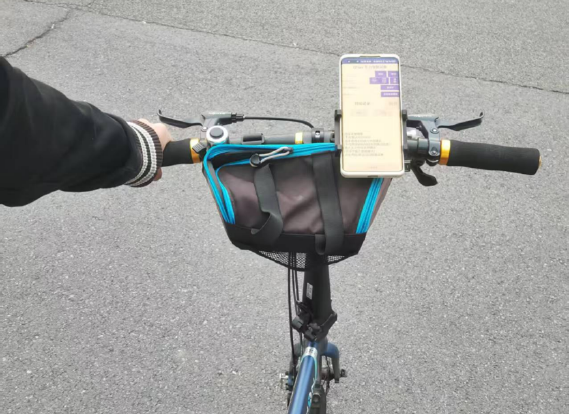}
\caption{Smartphone mounted on the bicycle handlebar for cycling data collection.}
\label{fig:device}
\end{figure}

\textbf{Data Collection:} Cycling data are collected using a self-developed smartphone application that accesses the built-in IMU, GNSS, magnetometer, and Wi-Fi sensors. During each session, as illustrated in Fig.~\ref{fig:device}, the smartphone is securely mounted on the bicycle handlebar using a phone holder, and the application continuously records multi-modal sensor streams throughout the experiment. To ensure the diversity of motion patterns and sensor characteristics, the data are acquired by multiple participants riding different types of bicycles over various road conditions.

\textbf{Data Processing:} The collected raw data are subsequently processed into training-ready samples. Sensor streams from different modalities are first temporally aligned through interpolation to a unified time base. A GNSS/INS global pose graph optimization is then performed by jointly modeling IMU pre-integration factors and GNSS position factors, which provides smoothed trajectories and per-epoch ground-truth velocities. These optimized estimates serve as the supervisory signals for neural network training.

\textbf{Neural Network Training:} A lightweight inertial odometry network based on the Mixture of Experts (MoE) is trained offline on the processed dataset. The network takes a sequence of IMU measurements and orientation in the body frame as input, and predicts the three-dimensional body-frame velocity at the end of the window together with the associated covariance matrix. The MoE design allows the network to capture diverse cycling motion patterns while maintaining a low computational cost.

\begin{figure}[!t]
\centering
\includegraphics[width=\columnwidth]{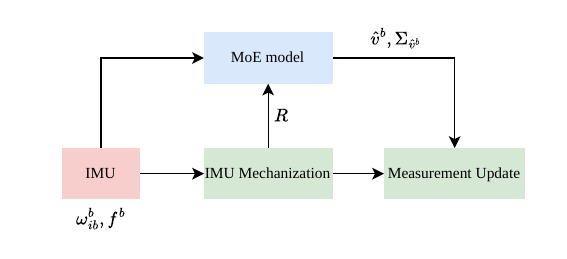}
\caption{Framework of the error-state EKF that tightly couples raw IMU measurements with the velocity predictions of the MoE network.}
\label{fig:filter_inference}
\end{figure}

\textbf{Filter Inference:} At inference time, as illustrated in Fig.~\ref{fig:filter_inference}, an error-state Extended Kalman Filter (EKF) tightly couples raw IMU measurements with the velocity predictions of the MoE network, which are produced every 0.1 seconds. Specifically, the filter performs mechanization using raw IMU data to obtain prior estimates of orientation, velocity, and position, and then applies measurement updates using the MoE output to jointly estimate orientation, velocity, position, and IMU biases.

Within this pipeline, the IMU measurements are utilized in two complementary ways. First, they drive the IMU mechanization inside the filter to produce prior state estimates. Second, together with the orientation provided by the filter, they are fed into the MoE network, which essentially treats IMU measurements as motion indices and maps IMU--orientation sequences to velocity through the learned correlations. This design avoids redundant use of the same signal while exploiting both the instantaneous dynamics captured by raw IMU data and the motion patterns learned from historical observations.

\section{Algorithm Description}
\subsection{MoE Model}
\begin{figure*}[!t]
\centering
\includegraphics[width=\textwidth]{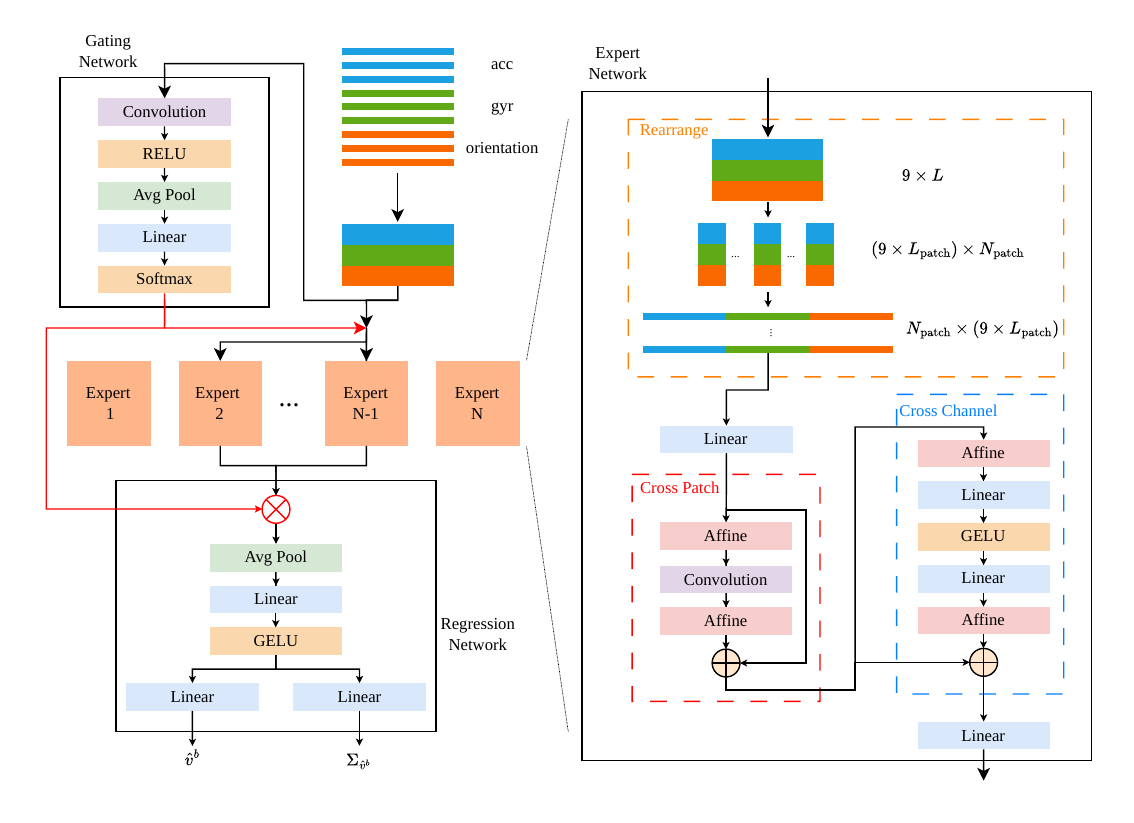}
\caption{Algorithm framework of the MoE model.}
\label{fig:main_algorithm}
\end{figure*}
This section introduces the MoE model proposed in the algorithm and Fig.~\ref{fig:main_algorithm} illustrates the framework of the MoE model.

\subsubsection{Network Architecture}
The architecture of the MoE model is composed of three main components: the gating network, the expert networks and the regression network, which are described separately below.

In the MoE framework, the gating network determines a small subset of K experts to be activated for each input sample. This selective activation preserves the expressive capacity of the model while significantly reducing computational overhead, making the gating mechanism a critical component. To further improve efficiency, the experts are activated according to a Top-K selection rule. Specifically, for each batch of input, the gating network ranks the expert scores and selects the top K experts with the highest weights for inference, while discarding the rest.

The gating network is implemented as follows: each sample passes sequentially through a convolutional layer, RELU layer, a global average pooling layer, and a linear layer to generate expert-specific weights. These weights are normalized via a softmax layer to form a probability distribution. For each sample, the top K experts with the highest probabilities are chosen, their probabilities are renormalized, and a sparse one-hot gating matrix is constructed to allocate samples to the corresponding experts.

The expert network adopts ResMLP~\cite{touvron2022resmlp} as the backbone for feature extraction, which is also employed by LLIO-Net~\cite{wang2022llio}. The design consists of three modules: a feature convert module, a ResMLP module, and a regression module. The feature convert module reorganizes the raw input into feature matrices, the ResMLP module extracts high-level representations, and the regression module integrates features from the Top-K experts to predict velocity and the associated covariance.

The input to the feature extraction module consists of raw IMU measurements and the corresponding orientations from time $t-L$ to $t$, forming a $9 \times L$ matrix. In the feature convert module, the input is divided into $N_{\text{patch}}$ patches, each containing $L_{\text{feature}}$ measurements, where $L = N_{\text{patch}} \times L_{\text{feature}}$. Each patch is then flattened and all the features are concatenated to obtain $N_{\text{patch}}$ embeddings of dimension $(9 \times L_{\text{feature}})$. These $N_{\text{patch}}$ embeddings are then fed into the ResMLP module.

The ResMLP module consists of two components: a cross-patch interaction module and a cross-channel interaction module. Before entering the ResMLP, a linear layer maps the input from dimension $N_{\text{patch}} \times (9 \times L_{\text{feature}})$ to $N_{\text{patch}} \times D$.

In the cross-patch interaction module, the input is sequentially passed through an affine layer, a one-dimensional convolutional layer, a linear layer, and another affine layer. Here, the affine layer serves as an alternative to layer normalization, performing element-wise scaling and shifting on the input. The affine operation is defined as:
\begin{equation}
\mathrm{AFF}_{\alpha,\beta}(x) = \operatorname{Diag}(\alpha)\,x + \beta
\end{equation}
where $\alpha$ and $\beta$ are learnable parameters. It should be noted that when $\text{AFF}(\cdot)$ is applied to a matrix, the operation is performed independently on each column. To facilitate multilayer stacking, both the convolutional and linear layers preserve the input dimension, which requires the convolution kernel size to be set to $1$.

In the cross-channel interaction module, the data sequentially pass through a linear layer, an activation function layer, and another linear layer. The second linear layer maps the representation from $N_{\text{patch}} \times D$ to $N_{\text{patch}} \times D_{\text{out}}$.

The regression network concatenates the weighted features extracted by the Top-K experts and then fuses them through a linear layer. This fused representation is further processed by a global average pooling layer, followed by a linear layer and an activation function layer for feature transformation. Finally, two separate linear layers are employed to estimate two three-dimensional vectors: the velocity and the diagonal elements of the covariance matrix.

\subsubsection{Training Methodology}
\begin{figure}[!t]
\centering
\includegraphics[width=\columnwidth]{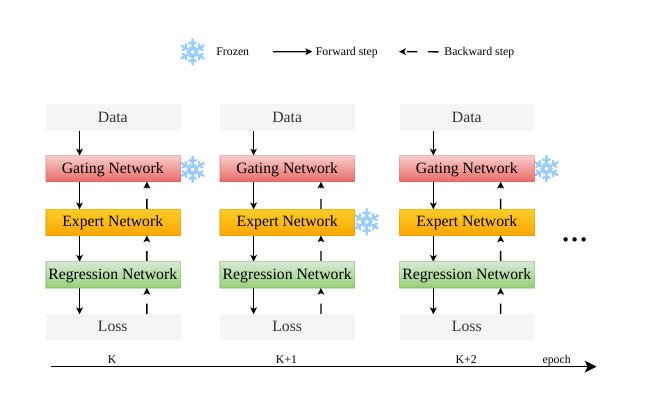}
\caption{The alternating parameter-freezing training strategy.}
\label{fig:train_algorithm}
\end{figure}
As illustrated in Fig.~\ref{fig:train_algorithm}, to improve the overall performance of the MoE network, we adopt an alternating parameter-freezing training strategy. Specifically, at epoch $K$, the parameters of the gating network are frozen, meaning that no gradients are computed for the gating network during backpropagation and its parameters are not updated. In the subsequent epoch $K+1$, the gating network is unfrozen while the parameters of all expert networks are frozen. This alternating process is performed every epoch.

By decoupling the optimization of the gating network and the expert networks, the proposed strategy enables the gating network to learn more accurate routing decisions, while allowing the expert networks to better specialize in processing the assigned samples. As a result, both routing quality and expert performance are improved.

Beyond the alternating training strategy, a capacity constraint is further introduced to ensure balanced expert utilization. In the MoE architecture, if all inputs are routed to only a few experts, the training efficiency will be greatly reduced, and the model will gradually degenerate from a Mixture of Experts into an equivalent single-expert network. This occurs because when the gating network tends to select a few fixed experts, the corresponding weight vectors of these experts increase during backpropagation, making it difficult for the other experts to contribute.

To mitigate this issue, we introduce an \emph{expert capacity} ($\mathrm{Capacity}$) constraint in the gating network to optimize load balancing among experts. The expert capacity is defined as:
\begin{equation}
\mathrm{Capacity}_i = \left\lceil \frac{c \times B}{N} \right\rceil
\end{equation}
where $\lceil \cdot \rceil$ denotes the ceiling operation. This formulation ensures that, within a batch of data, each expert can receive at most $\frac{c \times B}{N}$ inputs. If the number of inputs assigned to a certain expert exceeds this limit, the excess inputs are reassigned to the expert with the next highest weight. If that expert also exceeds its capacity, the input is passed further down to subsequent experts in descending order of their weights.

\subsubsection{Training Loss}
The training of the MoE model is primarily based on three loss functions:
\begin{itemize}
    \item \textbf{Mean Squared Error (MSE) Loss}: applied to the predicted velocity $\hat{v}^b$.
    \item \textbf{Negative Log-Likelihood (NLL) Loss}: applied jointly to $\hat{v}^b$ and its covariance $\Sigma_{\hat{v}^b}$.
    \item \textbf{Auxiliary Balancing Loss}: used to encourage balanced utilization across experts.
\end{itemize}

The mean squared error (MSE) loss is defined as:
\begin{equation}
\mathcal{L}_{MSE}\left(v, \hat{v}\right) = \frac{1}{n} \sum \left\|v -\hat{v}\right\|^2
\end{equation}
where $v^b$ denotes the ground truth velocity in the body frame, and $\hat{v}^b$ is the velocity predicted by the MoE model. Through minimizing $\mathcal{L}_{MSE}$, the MoE model learns to predict the three-dimensional velocity.

The negative log-likelihood (NLL) loss is defined as:
\begin{equation}
\mathcal{L}_{NLL}\left(v, \hat{v}, \hat{\Sigma}\right) =
\frac{1}{n} \sum \Bigg(
\frac{1}{2} \log \det \hat{\Sigma} +
\frac{1}{2} \|v - \hat{v}\|_{\hat{\Sigma}}^2
\Bigg)
\end{equation}
where $\hat{\Sigma}_{\hat{v}^b}$ denotes the covariance matrix corresponding to the predicted velocity $\hat{v}^b$. The Mahalanobis distance $\|v - \hat{v}\|_{\hat{\Sigma}}^2$ is defined as:
\begin{equation}
 \|v - \hat{v}\|_{\hat{\Sigma}}^2 =
 \left( v - \hat{v} \right)^T \hat{\Sigma} \left( v - \hat{v} \right)
\end{equation}
Through minimizing $\mathcal{L}_{NLL}$, the MoE model learns to predict the three-dimensional velocity along with its covariance.

The auxiliary balancing function is designed to encourage a balanced allocation of outputs across experts. It is implemented as follows: for a batch of $B$ inputs, the importance of expert $i$, denoted as $I_i$, is defined as the sum of the expert weight vectors received by this expert over the entire batch:
\begin{equation}
I_i = \sum_{t=1}^{B} g_i(x_t)
\end{equation}
where $g_i(x)$ denotes the gating weight generated by the gating network for input $x$ with respect to expert $i$. Ideally, the importance of each expert should be uniformly distributed across all experts, i.e.,
\begin{equation}
I_i \approx \frac{1}{N} \sum_{j=1}^{N} I_j
\end{equation}

Similarly, in addition to considering the continuous probability distribution, the actual number of times each expert is selected should also be taken into account. The load of expert $i$, denoted as $L_i$, is defined as:
\begin{equation}
L_i = \sum_{t=1}^{B} \mathbf{I}\!\left(i \in S(x_t)\right)
\end{equation}
where $\mathbf{I}(\cdot)$ is the indicator function, and $S(x)$ denotes the set of experts assigned to input $x$. Ideally, the load distribution across experts should be approximately uniform, i.e.,
\begin{equation}
L_i \approx \frac{B}{N}
\end{equation}
Based on the above, to encourage balanced allocation of input data among experts, the importance loss and load loss can be formulated as follows:
\begin{equation}
\begin{aligned}
\mathcal{L}_{\text{importance}} &= \sum_{i=1}^{N} \left( \frac{I_i}{\sum_{j=1}^{N} I_j} - \frac{1}{N} \right)^2 \\
\mathcal{L}_{\text{load}} &= \sum_{i=1}^{N} \left( \frac{L_i}{\sum_{j=1}^{N} L_j} - \frac{1}{N} \right)^2
\end{aligned}
\end{equation}

The importance loss measures the deviation of the experts' importance distribution from a uniform distribution. After normalizing the importance of each expert, it is compared with the ideal uniform value $1/N$, and larger deviations incur higher loss. The load loss encourages the number of times each expert is selected to be close to the mean, preventing certain experts from being over-utilized. The overall auxiliary balancing loss combines these two components as follows:
\begin{equation}
\mathcal{L}_{aux} = \mathcal{L}_{importance} + \mathcal{L}_{load}
\end{equation}

By minimizing $\mathcal{L}_{aux}$, the importance and load distributions are encouraged to be close to uniform. This prevents certain experts from being over-utilized or under-utilized, leading to a more balanced distribution of importance and load among experts. Consequently, the MoE model can effectively leverage the diverse expertise of all experts while maintaining computational efficiency and training stability.

During the training process, the MoE model is first trained by minimizing $\mathcal{L}_{MSE} + \lambda\,\mathcal{L}_{aux}$ until convergence. Subsequently, $\hat{v}_t^b$ and $\Sigma_{\hat{v}^b}$ are jointly trained by minimizing $\mathcal{L}_{NLL} + \lambda\,\mathcal{L}_{aux}$ until convergence, where $\lambda$ is the weighting coefficient for the auxiliary balancing loss.

\subsection{Extended Kalman Filter}
The MoE model predicts velocity in the body frame, enabling direct constraints on the current frame without the need to track historical states, thus simplifying the filter architecture.

\subsubsection{System State Definition}
The system state of the Extended Kalman Filter (EKF) is defined as:
\begin{equation}
X_t = \left( R_t, \; v_t^n, \; p_t^n, \; b_{g,t}, \; b_{a,t} \right)
\end{equation}
where $b_{g,t}$ and $b_{a,t}$ denote the biases of the gyroscope and accelerometer, respectively. The Kalman filter in this work is formulated based on the error of the system state, which is defined as:
\begin{equation}
\delta X_t = \left( \phi_t, \; \delta v_t^n, \; \delta p_t^n, \; \delta b_{g,t}, \; \delta b_{a,t} \right)
\end{equation}
where $\phi_t$ is a $3 \times 1$ vector representing the attitude error, defined as:
\begin{equation}
\phi_t = \log_{SO(3)}\!\left( R_t \hat{R}_t^{-1} \right)
\end{equation}
where $\log_{SO(3)}$ denotes the logarithmic map from the Lie group $\mathrm{SO}(3)$ to its Lie algebra $\mathfrak{so}(3)$.

\subsubsection{State Propagation}
The filter employs a simplified kinematic formulation for state propagation:
\begin{equation}
\renewcommand{\arraystretch}{1.4}
\begin{array}{l}
\hat{R}_{t+1} = \hat{R}_t \exp_{SO(3)}\!\left(\omega_{ib,t}^b - \hat{b}_{g,t}\right) \\
\hat{v}_{t+1}^n = \hat{v}_t^n + g\,\Delta t + R_t f_t^b\,\Delta t - \hat{b}_{a,t}\,\Delta t \\
\hat{p}_{t+1}^n = \hat{p}_t^n + 0.5\left(\hat{v}_t^n + \hat{v}_{t+1}^n\right)\Delta t \\
\hat{b}_{g,t+1} = \hat{b}_{g,t} \qquad
\hat{b}_{a,t+1} = \hat{b}_{a,t}
\end{array}
\end{equation}

Here, $\exp_{SO(3)}$ denotes the exponential map from the Lie algebra $\mathfrak{so}(3)$ to the Lie group $\mathrm{SO}(3)$. The linearized propagation model of the error state is given by:
\begin{equation}
\delta\hat{X}_{t+1} = A_t\,\delta\hat{X}_t + B_t n_t
\end{equation}
where
$
n_t =
\begin{pmatrix}
n_{g,t} & n_{a,t} & \eta_{bg,t} & \eta_{ba,t}
\end{pmatrix}^T
$
represents the noise of the gyroscope and accelerometer, as well as the random walk noise of their biases. The corresponding covariance propagation model is given by:
\begin{equation}
P_{t+1} = A_t P_t A_t^{T} + B_t Q_t B_t^{T}
\end{equation}
where $P_t$ and $Q_t$ denote the covariance matrices of the system state and the noise at time $t$, respectively.

\subsubsection{Measurement Update}
The measurement update is as follows:
\begin{equation}
h(X_t) = \hat{R}_t^T v_t^n = \hat{v}_t^b + n_{\hat{v}_t^b}
\end{equation}
where $n_{\hat{v}_t^b}$ follows a Gaussian distribution $\mathcal{N}\!\left(0, \Sigma_{\hat{v}_t^b}\right)$. The function $h(X_t)$ has nonzero partial derivatives only with respect to $\hat{\phi}_t$ and $\delta v_t^n$. The corresponding Jacobian matrix is given by:
\begin{equation}
\begin{aligned}
H\!\left(\hat{\phi}_t\right)
&= \frac{\partial h(X)}{\partial \hat{\phi}_t}
= \hat{R}_t^T \left(\hat{v}_t^n\right)_\times \\[4pt]
H\!\left(\delta\hat{v}_t^n\right)
&= \frac{\partial h(X)}{\partial \delta\hat{v}_t^n}
= \hat{R}_t^T
\end{aligned}
\end{equation}

Finally, the Kalman gain and the update can be computed as follows:
\begin{equation}
\begin{aligned}
K_t &= P_t H_t^T \left( H_t P_t H_t^T + \Sigma_{\hat{v}_t^b} \right)^{-1} \\[4pt]
\hat{X}_t &\gets \hat{X}_t \oplus \left( K_t \left( h(X_t) - \hat{v}_t^b \right) \right) \\[4pt]
P_t &= \left( I - K_t H_t \right) P_t \left( I - K_t H_t \right)^{T}
       + K_t\,\Sigma_{\hat{v}_t^b}\,K_t^{T}
\end{aligned}
\end{equation}

Here, $\oplus$ denotes the operation of adding the error state to the system state. It should be noted that the attitude update is computed as $\hat{R}_t = \exp_{SO(3)}\!\left(\hat{\phi}_t\right)\,\hat{R}_t^{-}$.

\section{Experiments}
\subsection{Setup}
\subsubsection{Dataset}
Following the data collection and processing pipeline described in Section~\ref{sec:system_overview}, the complete dataset comprises more than 25 hours of cycling sensor recordings covering a wide range of road conditions. The data were collected by eight participants riding bicycles of different models, yielding diverse motion patterns and IMU systematic errors. The dataset is partitioned into 70\% for training, 10\% for validation, and 20\% for testing.

\subsubsection{Evaluation Metrics}
To evaluate the localization performance of different methods, we define the following metrics similar to TLIO.

The \textbf{Absolute Translation Error (ATE, m)} measures the root-mean-square error (RMSE) between the estimated and ground-truth trajectories, expressed in meters, and is defined as:
\begin{equation}
\mathrm{ATE} = \sqrt{\frac{1}{n} \sum_{t=1}^{n} \left\| p_t^{n} - \hat{p}_t^{n} \right\|^2 }
\end{equation}
where $p_t^{n}$ and $\hat{p}_t^{n}$ denote the ground-truth and estimated positions at time $t$, respectively.

The \textbf{Relative Translation Error (RTE, m)} evaluates the relative trajectory error over a given time interval in a locally gravity-aligned coordinate frame, and is therefore insensitive to accumulated heading errors. It is defined as:
\begin{equation}
\mathrm{RTE} = \sqrt{\frac{1}{n} \sum_{t=1}^{n} \left\| R_{\text{yaw}}^{T} \left( p_t^{n} - p_{t-\Delta t}^{n} \right)
- \hat{R}_{\text{yaw}}^{T} \left( \hat{p}_t^{n} - \hat{p}_{t-\Delta t}^{n} \right) \right\|^2 }
\end{equation}
where $R_{\text{yaw}}$ and $\hat{R}_{\text{yaw}}$ denote the yaw rotation matrices corresponding to the ground-truth and estimated trajectories, respectively. In our evaluation, one-minute segments are adopted, i.e., $\Delta t = 1~\text{minute}$.

\subsubsection{Evaluation Protocol}
To simulate realistic GNSS outage scenarios, all methods are evaluated under a standardized outage protocol. Each test sequence is segmented into repeating cycles: the filter is first provided with 10 seconds of reference position updates for initialization, after which position corrections are suspended for 120 seconds to simulate a GNSS outage. At the end of each outage window, another 10-second initialization phase is applied before the next outage begins. Performance metrics are computed exclusively over the 120-second outage intervals, so that ATE and RTE reflect pure inertial dead-reckoning accuracy without the benefit of external position aiding.

\begin{figure*}[!t]
\centering
\includegraphics[width=\textwidth]{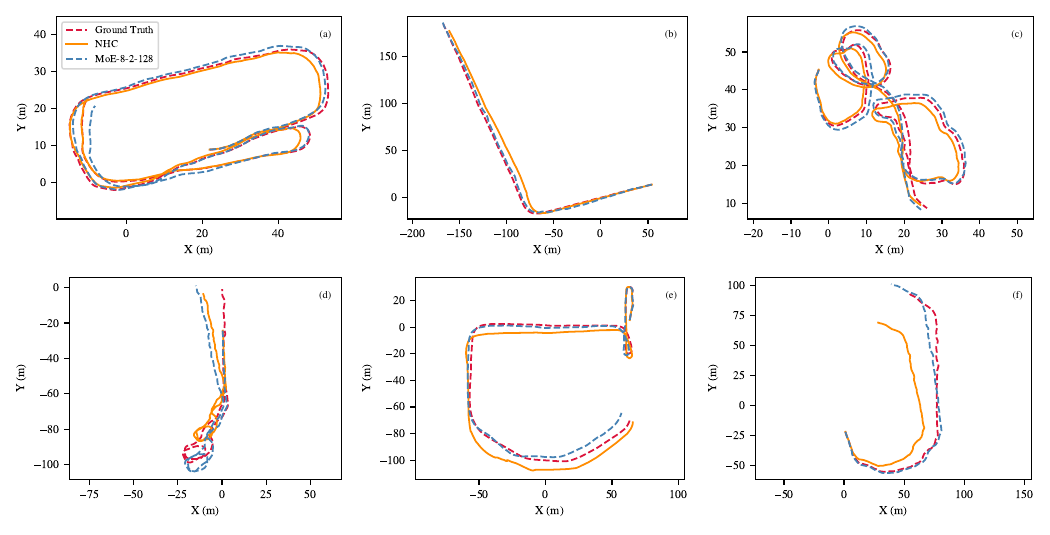}
\caption{Estimated horizontal trajectories of MoE-$8$-$2$-$128$ and the NHC baseline against the ground truth on six representative test sequences. The top row shows three paved-road sequences, while the bottom row shows three unpaved-road sequences, on which the NHC baseline exhibits markedly larger drift.}
\label{fig:traj_moe_vs_nhc}
\end{figure*}

\subsection{Comparison with the Model-based Baseline}
Throughout the experiments, we adopt the naming convention MoE-$N$-$K$-$D$, where $N$ is the size of the expert pool, $K$ is the per-sample activation budget under top-$K$ routing, and $D$ is the embedding dimension shared by the experts and the gating network; the learned baselines introduced later (ResMLP and ResNet) follow a similar convention, with the trailing number denoting the embedding dimension $D$.

We first contrast MoE-$8$-$2$-$128$ with a representative model-based baseline, the nonholonomic-constraint (NHC) filter described in Section~II-A, in order to establish whether classical kinematic constraints remain a viable reference for cycling dead reckoning. The NHC implementation follows the canonical formulation for two-wheeled platforms: an error-state EKF that integrates IMU mechanization in the propagation step and, in the update step, enforces zero lateral and vertical velocity in the vehicle frame, with the IMU-to-vehicle lever arm and installation angle augmented into the state and online-estimated. During the simulated GNSS outages, no external position aiding is provided to either method, so the comparison reflects pure inertial dead-reckoning behavior under identical conditions.

\begin{table}[!t]
\centering
\caption{Comparison between MoE-$8$-$2$-$128$ and the NHC Baseline on the Overall Test Set and on the Paved/Unpaved-Road Splits\label{tab:moe_vs_nhc}}
\setlength{\tabcolsep}{3pt}
\scriptsize
\begin{tabular}{lcccccc}
\hline
            & \multicolumn{2}{c}{Overall} & \multicolumn{2}{c}{Paved road} & \multicolumn{2}{c}{Unpaved road} \\
Method       & ATE (m) & RTE (m) & ATE (m) & RTE (m) & ATE (m) & RTE (m) \\ \hline
NHC         & 10.82     & 4.01     & \textbf{8.54}     & 3.36     & 13.86     & 4.44     \\
MoE-8-2-128 & \textbf{9.49} & \textbf{2.58} & 9.86 & \textbf{2.44} & \textbf{8.76} & \textbf{2.86} \\
\hline
\end{tabular}
\end{table}

Table~\ref{tab:moe_vs_nhc} summarizes the result. Aggregated over the full test set, the NHC baseline trails MoE-$8$-$2$-$128$ on both ATE and RTE, indicating that on average the rigid no-side-slip and no-vertical-velocity assumptions cannot fully capture the lateral leaning, frame compliance, and small vertical excursions of real bicycle motion. Resolving this aggregate trend along road surface, however, reveals a clearly bimodal behavior: on the \textit{Paved road} split, NHC is in fact competitive, slightly outperforming MoE on ATE and remaining within $0.92$\,m on RTE, since flat asphalt largely satisfies its kinematic assumptions and the online lever-arm/installation-angle estimation can stabilize the filter. On the \textit{Unpaved road} split, by contrast, NHC degrades sharply, whereas MoE remains close to its overall accuracy. This is consistent with the analysis in Section~II-A: on unpaved surfaces, frequent micro-vibrations and irregular wheel--ground interactions repeatedly violate the constraint assumptions, injecting persistent bias into the lateral/vertical pseudo-measurements and ultimately driving the position estimate off, while the data-driven predictor has been exposed to similar conditions during training and produces a velocity estimate that absorbs these surface-induced disturbances rather than being misled by them.

Fig.~\ref{fig:traj_moe_vs_nhc} illustrates this contrast qualitatively on six representative sequences. On the three paved-road sequences in the top row, the NHC and MoE trajectories both stay close to the ground truth, in agreement with their comparable error figures on the \textit{Paved road} split. On the three unpaved-road sequences in the bottom row, by contrast, the NHC trajectories deviate noticeably from the ground truth, whereas MoE continues to track it closely, reflecting the accumulation of model error in NHC under surface-induced constraint violations. Together, the table and the trajectory comparison motivate the remainder of this section, which focuses on benchmarking MoE-$8$-$2$-$128$ against the strongest learned alternatives, with the understanding that the model-based NHC pipeline---while competitive on well-behaved paved surfaces---is no longer a reliable baseline under the more challenging road conditions that cycling localization actually has to handle.
\subsection{Comparison with Learned Baselines}

\begin{table}[!t]
\centering
\caption{Performance Comparison of MoE, ResMLP, and ResNet Models with Multiple Hyperparameter Settings\label{tab:model_comparision}}
\setlength{\tabcolsep}{9pt}
\footnotesize
\begin{tabular}{lccccc}
\hline
Model       & FLOPs & IE    & ATE & RTE & Params \\
            & (M)   & (m/s) & (m) & (m) & (M)    \\ \hline
MoE-8-2-64  & \textbf{9.35}   & 0.353 & 10.29& 3.05 & 2.08 \\
MoE-8-2-128 & 28.76  & 0.333 & 9.49 & 2.58 & 7.18 \\
MoE-8-2-256 & 104.33 & \textbf{0.327} & \textbf{9.15} & \textbf{2.52} & 27.00 \\
ResMLP-128  & 12.63  & 0.389 & 10.71& 3.01 & \textbf{0.87} \\
ResMLP-256  & 49.44  & 0.350 & 9.73 & 2.72 & 3.42 \\
ResMLP-512  & 195.64 & 0.349 & 9.91 & 2.56 & 13.52 \\
ResNet-128  & 17.03  & 0.386 & 10.99& 2.78 & 1.17 \\
ResNet-256  & 67.26  & 0.365 & 10.63& 3.05 & 4.60 \\
ResNet-512  & 267.26 & 0.352 & 10.38& 2.76 & 18.24 \\
\hline
\end{tabular}
\end{table}

Table~\ref{tab:model_comparision} presents an overall comparison among three network architectures: the proposed MoE model, the ResMLP backbone (used in LLIO), and the ResNet-based model (used in TLIO). For each architecture, three hyperparameter configurations are evaluated, jointly considering computational cost, model size, and inference accuracy so as to ensure a fair comparison across different design trade-offs. In addition to ATE and RTE, we report the inference error (IE), defined as $\frac{1}{n} \sqrt{\sum \| \mathbf{v}_t^n - \hat{\mathbf{v}}_t^n \|^2}$, which directly measures the prediction accuracy of the network outputs and differs slightly from the standard root-mean-square error (RMSE).

Across all hyperparameter settings, the proposed MoE model attains accuracy comparable to or better than both ResMLP and ResNet while requiring substantially less computation. In particular, MoE-8-2-128 reaches an IE of $0.333$\,m/s and an ATE of $9.49$\,m at only $28.76$\,M FLOPs, whereas ResMLP-512 and ResNet-512 consume $195.64$\,M and $267.26$\,M FLOPs (roughly $7\times$ and $9\times$ more) yet achieve worse or comparable accuracy. This confirms that the MoE design offers a more favorable accuracy and efficiency trade-off than single-network backbones of equivalent representational capacity.

To complement these aggregate statistics with a more direct view of the localization behavior, we select MoE-8-2-128, ResMLP-512, and ResNet-512 as representative baselines and compare them at both the trajectory and the error-distribution levels. The latter two are the highest-capacity variants of the ResMLP and ResNet families and therefore correspond to the strongest single-network competitors against the proposed MoE model.

\begin{figure*}[!t]
\centering
\includegraphics[width=\textwidth]{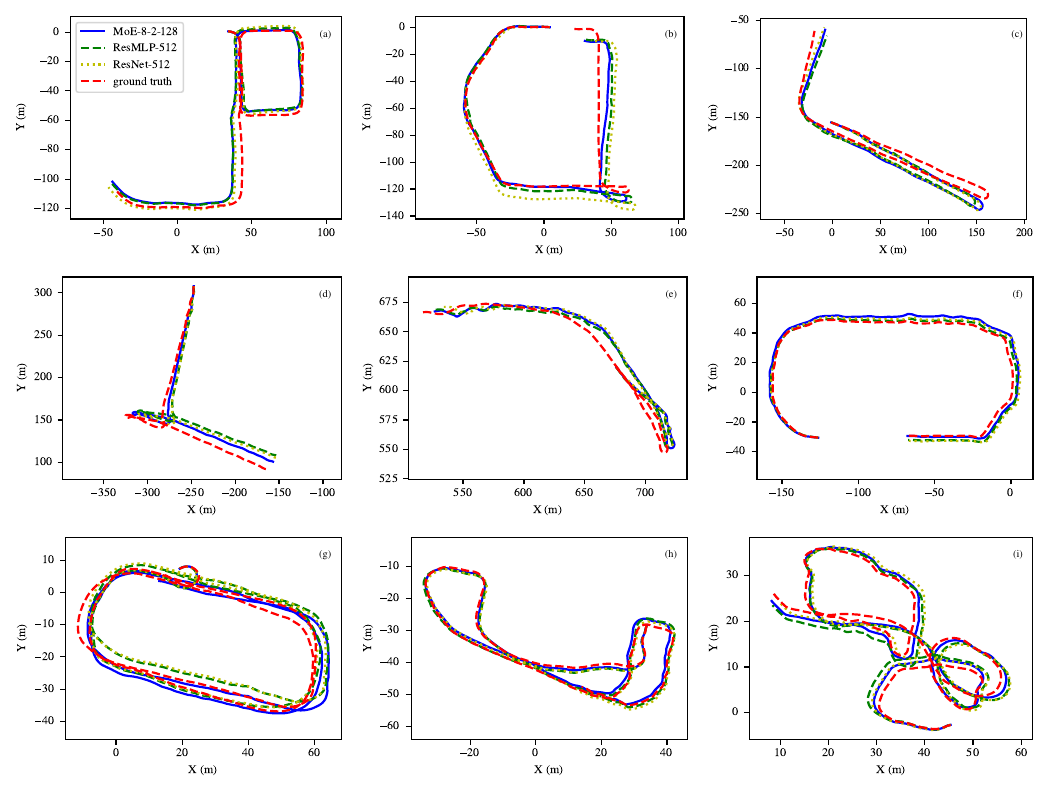}
\caption{Estimated horizontal trajectories of MoE-8-2-128, ResMLP-512, and ResNet-512 against the ground truth on nine representative test sequences covering different riders, bicycles, and road conditions. Subfigures (a)--(f): longer paved-road sequences with extended straight and curved segments; subfigures (g)--(i): looped and geometrically complex trajectories with tight curves and multi-turn motion.}
\label{fig:combined_traj}
\end{figure*}

Fig.~\ref{fig:combined_traj} overlays the estimated horizontal trajectories of the three baselines and the ground truth on nine representative test sequences, spanning different riders, bicycles, and road conditions. Subfigures (a)--(f) correspond to longer sequences with extended straight and curved segments, while subfigures (g)--(i) contain looped and geometrically complex trajectories featuring tight curves and multi-turn motion. Across all sequences, the MoE-8-2-128 trajectory remains closest to the ground truth, with smaller lateral offsets along long segments and tighter alignment through sharp turns and loops. ResNet-512 generally exhibits the largest deviations while ResMLP-512 falls in between, consistent with the trend observed in Table~\ref{tab:model_comparision}.

\begin{figure}[!t]
\centering
\includegraphics[width=\columnwidth]{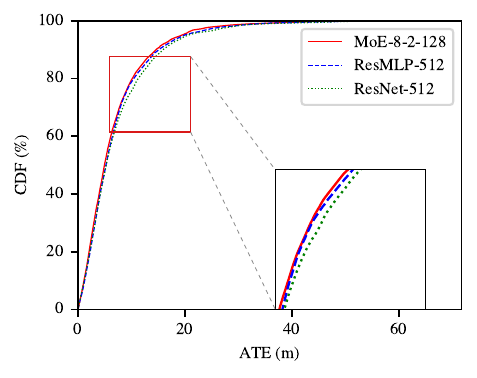}\\
\includegraphics[width=\columnwidth]{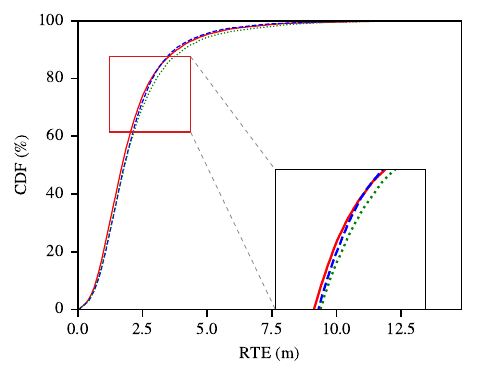}
\caption{Cumulative distribution functions of per-sequence ATE (top) and RTE (bottom) on the test set for MoE-8-2-128, ResMLP-512, and ResNet-512.}
\label{fig:cdf}
\end{figure}

To quantify this comparison over the full test set rather than a few selected sequences, Fig.~\ref{fig:cdf} reports the cumulative distribution functions (CDFs) of the per-sequence ATE and RTE for the same three models. In a CDF plot, a curve that lies further to the upper-left indicates that more sequences fall below a given error level. Both CDFs show that MoE-8-2-128 lies on or slightly above the ResMLP-512 and ResNet-512 curves over the bulk of the error range, most clearly between the $60\%$ and $90\%$ quantiles, as highlighted in the zoomed inset. This means that for a typical sequence the proposed model attains a higher cumulative probability at a given error threshold and therefore a smaller error. Combined with the trajectory comparison in Fig.~\ref{fig:combined_traj}, these results show that the proposed MoE model delivers more accurate localization on typical sequences than the strongest ResMLP and ResNet variants, while operating at a substantially lower computational cost.

\subsection{Robustness and Generalization Analysis}

To further assess robustness under realistic distribution shifts, we partition the test set along three orthogonal dimensions (rider, bicycle, and road surface) into six splits: \textit{People seen}/\textit{unseen}, \textit{Bicycle seen}/\textit{unseen}, and \textit{Paved}/\textit{Unpaved road}. The \textit{unseen} splits contain participants or bicycles absent from training and therefore probe out-of-distribution generalization.

\begin{table*}[!t]
\centering
\caption{Performance Comparison across Different Evaluation Settings\label{tab:scenario_analysis}}
\setlength{\tabcolsep}{8pt}
\begin{tabular}{lccccccccccccc}
\hline
 & & \multicolumn{2}{c}{People seen} & \multicolumn{2}{c}{People unseen} & \multicolumn{2}{c}{Bicycle seen} & \multicolumn{2}{c}{Bicycle unseen} & \multicolumn{2}{c}{Paved road} & \multicolumn{2}{c}{Unpaved road} \\
Model & FLOPs (M)
& ATE & RTE
& ATE & RTE
& ATE & RTE
& ATE & RTE
& ATE & RTE
& ATE & RTE \\
\hline
MoE-8-2-64  & \textbf{9.35}   & 9.12 & 3.92 & 10.81 & 2.22 &  9.52 & 2.08 &  8.53 & \textbf{3.19} & 10.98 & 3.15 & 8.66 & 2.75 \\
MoE-8-2-128 & 28.76  & 6.58 & \textbf{2.33} & 10.75 & \textbf{2.02} &  \textbf{7.87} & \textbf{1.67} &  \textbf{8.15} & 3.20 &  9.86 & 2.44 & 8.76 & 2.86 \\
MoE-8-2-256 & 104.33 & \textbf{5.86} & 2.38 & \textbf{10.41} & 2.16 &  8.64 & 1.75 &  8.31 & 3.49 &  \textbf{9.57} & \textbf{2.34} & \textbf{7.92} & \textbf{2.60} \\
ResMLP-128  & 12.63  & 8.69 & 2.98 & 11.75 & 2.67 & 10.07 & 2.13 & 12.16 & 4.74 & 10.15 & 2.43 & 11.86 & 3.61 \\
ResMLP-256  & 49.44  & 8.13 & 2.72 & 10.76 & 2.23 &  8.86 & 1.89 &  9.81 & 3.79 &  9.84 & 2.48 & 9.34 & 2.91 \\
ResMLP-512  & 195.64 & 6.98 & 2.42 & 11.20 & 2.20 &  9.39 & 1.93 &  9.51 & 3.35 & 10.24 & 2.41 & 8.69 & 2.68 \\
ResNet-128  & 17.03  & 11.89 & 3.86 & 12.38 & 2.87 & 10.53 & 2.15 & 10.37 & 3.98 & 11.05 & 2.43 & 11.09 & 3.13 \\
ResNet-256  & 67.26  & 8.99 & 3.46 & 11.71 & 2.56 & 10.11 & 2.12 & 9.88 & 4.13 & 11.10 & 2.91 & 9.09 & 2.78 \\
ResNet-512  & 267.26 & 6.87 & 2.43 & 10.94 & 2.03 & 9.56 & 1.97 & 10.77 & 3.46 & 10.62 & 2.66 & 8.93 & 2.77 \\

\hline
\end{tabular}
\end{table*}

Table~\ref{tab:scenario_analysis} reports the resulting ATE and RTE. Among the three dimensions, the bicycle dimension imposes the largest generalization gap, with every model degrading clearly on \textit{Bicycle unseen}; the rider dimension is the mildest (RTE even decreases on \textit{People unseen} for most models, suggesting that rider-induced variation is largely absorbed during training); and the road-surface dimension lies in between. On these challenging splits the MoE variants attain the best ATE and RTE in every case, with the largest margin on \textit{Bicycle unseen} (MoE keeps ATE within $8.15$ to $8.53$\,m while ResMLP-128 and ResNet-128 deteriorate to $12.16$ and $10.37$) and on \textit{Unpaved road} (MoE-8-2-256 leads all baselines on both metrics), indicating that the Mixture-of-Experts design degrades more gracefully under the hardest shifts.

\begin{figure}[!t]
\centering
\includegraphics[width=\columnwidth]{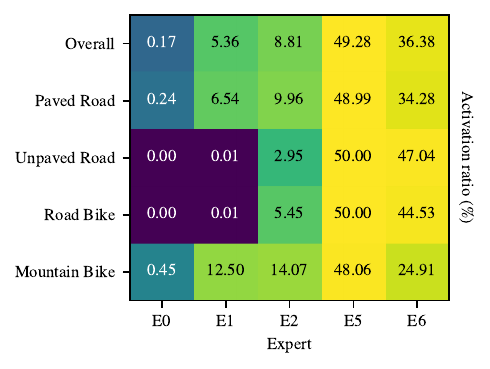}
\caption{Expert activation heatmap of MoE-8-2-128 under different road conditions and bicycle types. Each row corresponds to a subset of the test data and each column to an expert; cell values report the activation ratio (\%) of an expert within the corresponding subset. Three experts that are never activated by the gating network are omitted, leaving five active experts (E0, E1, E2, E5, E6).}
\label{fig:expert_heatmap}
\end{figure}

To clarify the source of this robustness, we take MoE-8-2-128 as a representative configuration and inspect its expert activation distribution in Fig.~\ref{fig:expert_heatmap}. The routing is highly sparse: three of the eight experts are never selected and are omitted, while E5 and E6 alone account for about $85\%$ of the activations. More importantly, the activation pattern shifts with the operating condition. On unpaved road, the share of E6 rises from $34.28\%$ to $47.04\%$ and E1/E2 are almost entirely suppressed, concentrating rough-surface samples on a high-capacity pair of experts. On the mountain bike, by contrast, routing becomes markedly more dispersed, with E1 and E2 jointly contributing $26.57\%$, reflecting the broader motion repertoire of mountain-bike riding. This condition-dependent specialization is the mechanism behind the gains observed on \textit{Bicycle unseen} and \textit{Unpaved road}: the gate reroutes challenging samples to experts that have learned the corresponding motion regimes, whereas a single-network backbone must share its full capacity across all conditions.

\subsection{Ablation Study}

We conduct two ablations on the key MoE hyper-parameters: the expert pool size $N$ and the per-sample activation budget $K$. The embedding dimension is fixed to $D=128$ throughout. We also report $N_{\text{act}}$, the number of experts actually exercised by the gating network on the test set, which is an indicator of how well the available expert pool is utilized.

\subsubsection{Effect of the Activation Budget $K$}

We fix $N=8$ and sweep $K\in\{1,2,3,8\}$, so that all configurations share the same parameter count ($7.18$\,M) and only the activated compute varies. The case $K{=}N{=}8$ corresponds to dense activation and serves as a non-sparse upper-cost reference.

\begin{table}[!t]
\centering
\caption{Effect of the Activation Budget $K$ ($N=8$, $D=128$).\label{tab:ablation_k}}
\setlength{\tabcolsep}{8pt}
\footnotesize
\begin{tabular}{cccccc}
\hline
$K$ & $N_{\text{act}}$ & IE   & ATE  & RTE  & FLOPs \\
    &                  & (m/s)& (m)  & (m)  & (M)   \\
\hline
1 & 4 & 0.356 & 9.48 & 2.59 & 15.48 \\
2 & 5 & 0.333 & 9.49 & 2.58 & 28.76 \\
3 & 6 & 0.334 & 9.28 & 2.69 & 42.04 \\
8 & 8 & 0.374 & 10.99 & 3.21 & 108.46\\
\hline
\end{tabular}
\end{table}

As shown in Table~\ref{tab:ablation_k}, $K{=}1$ already attains competitive accuracy at $15.48$\,M FLOPs, but the single activation slot offers little incentive for routing diversity and four of the eight experts collapse to a dead state ($N_{\text{act}}{=}4$). Raising $K$ to $2$ recruits a fifth expert and improves accuracy, while $K{=}3$ lowers ATE at the cost of higher RTE. Notably, the dense reference $K{=}N{=}8$ incurs $3.8\times$ the cost of $K{=}2$ yet attains the worst accuracy: forcing every expert to contribute on every sample averages out the learned specialization and degenerates the model into an over-smoothed single network. Sparse top-$K$ routing therefore acts as a useful inductive bias rather than only a compute-saving device, and $K{=}2$ best balances accuracy, efficiency, and routing diversity.

\subsubsection{Effect of the Expert Pool Size $N$}

We fix $K=2$ and sweep $N\in\{4,6,8,10,12\}$, so that the per-sample activated compute is held roughly constant ($\approx\!28.76$\,M FLOPs) while the total expert capacity grows.

\begin{table}[!t]
\centering
\caption{Effect of the Expert Pool Size $N$ ($K=2$, $D=128$).\label{tab:ablation_n}}
\setlength{\tabcolsep}{8pt}
\footnotesize
\begin{tabular}{cccccc}
\hline
$N$ & $N_{\text{act}}$ & IE   & ATE  & RTE  & Params \\
    &                  & (m/s)& (m)  & (m)  & (M)    \\
\hline
4  & 4 & 0.329 & 9.49 & 2.65 & 3.68 \\
6  & 5 & 0.331 & 9.47 & 2.64 & 5.43 \\
8  & 5 & 0.333 & 9.49 & 2.58 & 7.18 \\
10 & 5 & 0.330 & 9.51 & 2.51 & 8.92 \\
12 & 5 & 0.335 & 9.86 & 2.58 & 10.67 \\
\hline
\end{tabular}
\end{table}

Table~\ref{tab:ablation_n} shows that $N_{\text{act}}$ saturates at five for every $N\geq 6$: beyond this point, additional experts add parameters without contributing to inference, consistent with the heatmap analysis of Fig.~\ref{fig:expert_heatmap}. Accuracy correspondingly forms a plateau over $N\in\{6,8,10\}$, where IE, ATE, and RTE span only $0.003$\,m/s, $0.04$\,m, and $0.13$\,m, respectively. $N{=}4$ is too small to cover the diversity of cycling motion, while $N{=}12$ wastes roughly $3.5$\,M parameters on unused experts and dilutes the gating signal. The chosen $N{=}8$ thus lies safely inside the plateau, providing a comfortable margin above the saturation point without the overhead of larger pools. Together, the two ablations confirm that MoE-$8$-$2$-$128$ is well matched to the routing capacity the data can actually exploit.

\section{Conclusion}
This work applies learned inertial odometry to bicycle localization, replacing the hand-crafted CDR and NHC motion models with a data-driven predictor that captures real-world cycling dynamics directly from raw IMU measurements, tightly coupled with an extended Kalman filter for three-dimensional pose estimation. To support training, we collected a real-world cycling dataset of over 25 hours spanning eight riders, multiple bicycle models, and both paved and unpaved road surfaces, providing the motion and IMU error diversity required for generalization. The predictor is realized as a sparsely-gated Mixture of Experts (MoE) network, trained with an alternating parameter-freezing scheme and a per-expert capacity constraint. The proposed MoE-$8$-$2$-$128$ achieves an inference error of $0.333$\,m/s, an ATE of $9.49$\,m, and an RTE of $2.58$\,m at only $28.76$\,M FLOPs, roughly $7\times$ and $9\times$ less than the ResMLP and ResNet baselines of LLIO and TLIO at comparable or superior accuracy.

The model further attains the lowest ATE and RTE on the out-of-distribution splits (unseen riders and unseen bicycles) as well as on the challenging unpaved-road condition, demonstrating cross-condition robustness beyond what hand-crafted cycling models can deliver. Visualization of gating decisions and the $N$-$K$ ablation indicate that this gain stems from condition-dependent expert specialization rather than ensemble averaging.

Future work will explore two directions. First, more sophisticated load-balancing strategies will be investigated to ensure that all experts are evenly utilized, thereby fully exploiting the capacity of the expert pool. Second, knowledge distillation will be applied to transfer the learned representations into a more compact model, further reducing computational overhead while preserving accuracy.

\bibliographystyle{IEEEtran}
\bibliography{references}

\end{document}